# Method for Searching of an Optimal Scenario of Impact in Cognitive Maps during Information Operations Recognition

Oleh Dmytrenko[1], Dmitry Lande[1,2], Oleh Andriichuk[1,2]

[1]Institute for Information Recording of National Academy of Sciences of Ukraine,
Kyiv, Ukraine
[2]National Technical University of Ukraine "Igor Sikorsky Kyiv Polytechnic Institute",
Kyiv, Ukraine

dmytrenko.o@gmail.com, dwlande@gmail.com, andriichuk@ipri.kiev.ua

**Abstract.** In this article, we consider the problem of choosing the optimal scenario of the impact between nodes based on of the introduced criteria for the optimality of the impact. Two criteria for the optimality of the impact, which are called the force of impact and the speed of implementation of the scenario, are considered. To obtain a unique solution of the problem, a multi-criterial assessment of the received scenarios using the Pareto principle was applied. Based on the criteria of a force of impact and the speed of implementation of the scenario, the choice of the optimal scenario of impact was justified. The results and advantages of the proposed approach in comparison with the Kosko model are presented.

**Keywords:** Cognitive Map, Optimal Scenario of Impact, Pareto Principle, Algorithm of Accumulative Impact, Force of Impact, Rank Distribution, Information Operation Recognition

## 1 Introduction

In today's world, it is difficult to overestimate the impact of information on people. Recently, the number of information sources has increased significantly and, accordingly, their influence is also increased. Information operations [1] may be one of the negative manifestations of this effect.

During the recognition of information operations [1], decision support systems (DSS) are used to make recommendations. When building knowledge bases of DSSs it often encounters the problem of lack of knowledge for describing a subject domain, which is corresponded to an object of an informational operation. In this case, a cognitive map, which is built automatically based on the textual data that corresponding to the object of the information operation, can be an additional tool for building a knowledge base of the DSS. Such cognitive map is a network of key terms that influence on each other. Rank distribution of nodes according to the degree of their impact on each other makes it possible to reveal key and the most influential components of the subject domain.

Rank distribution is one of the methods of ordering objects either physical or informational. In the case of certain numerical value can be assigned to each object from the collection, the ranking problems become formally trivial, since objects can be ranked by the value [2]. For example, the introduction of weight coefficients, characterizing the power of impact, turned out to be the main direction of development of the cognitive approach for analyzing a situation [3].

A cognitive map is a directed graph in which the edges (and sometimes the nodes) are characterized by weighted factors. A cognitive map, like any graph, is defined by the adjacency matrix $W$ [4], comprised of elements $w_{ij}$ – representing weight values of the edges connecting the corresponding nodes $u_1, u_2, ..., u_n$. The nodes of the cognitive map correspond to certain concepts, and edges are the casual (causal-consequential) connections between the corresponding concepts. Weight values are also used to analyze well-structured situations, where the value of the impact in different paths between the two nodes is summed up. However, the difficulty is that, firstly, it is not always clear how to determine such a numerical value, and secondly, such numerical values may be many and not always clear criterion for choosing one of them. In other words, the most complex, poorly formalized part of the problem of ranking is the choice of criterion for which the object is attributed to numerical values (formalization of objects).

In this article, the value of impact is calculated as follows:

1. In order to calculate the force of impact of one node to another (the impact of $u_i$ on $u_j$), it is necessary to find all the simple paths that exist between these two nodes. To find all the simple paths between a pair of nodes ($u_i, u_j$), the algorithm presented in work [5] is used. Each simple path represents a certain scenario of impact $(u_i, u_j)_k$.

2. Having introduced the criteria, the scenario of impact can be considered optimal for: the force and speed of the implementation of the scenario.

The purpose of this article is to justify a choice the optimal scenario of impact according to the introduced criteria.

## 2    Methods and Models for Nodes Ranking

In this section a short survey of other methods that can be used for cognitive maps for ranking of nodes according to the degree of their impact on each other makes is presented.

In the impulse method [6], each node in a cognitive map is assigned a value $v_i(t)$ at each moment of discrete time $t = 0, 1, 2, ...$. The weight of an edge is positive ($w_{ij} > 0$) if an increase in the weight of node $u_i$ causes an increase in the weight of node $u_j$. Conversely, the weight of an edge has a negative value ($w_{ij} < 0$) if decreasing the weight of node $u_i$ results in a decrease in the weight of node $u_j$. The weight $w_{ij} = 0$ if nodes $u_i$ and $u_j$ are not related.

The problem is to define the final value of node $v_i(t \to \infty)$, or in some cases the rate of change over time. To define $v_i(t)$ it is necessary to define how the node's value changes depending on its initial value, values of neighboring nodes, and weights of relations.

The basic procedure of cognitive mapping analysis is determined by the rule of the impulse process changing which is described in detail in [6]. According to this rule, the value of each concept $v_i(t)$ changes at the moment of discrete time $t$ ($t = 0,1,2,\ldots$) by the following equation:

$$v_i(t+1) = v_i(t) + \sum_{j=1}^{n} w_{ij} p_j(t), t = 0,1,2,\ldots.$$

where $n$ is the number of nodes in the graph.

An impulse is defined by the following equation:

$$p_j(t) = v_i(t) - v_i(t-1), t>0.$$

While investigating cognitive maps, values $v_i(0)$, which correspond to the concepts of the directed graph, and the pulse values $p_i(0)$ are defined at the initial moment of time $t = 0$.

In the Kosko model [7, 8] an influence value is calculated as follows: the indirect influence (i.e., the indirect effect) of action $I_p$ of vertex $i$ on vertex $j$ through path $P$ that connects vertex $i$ to vertex $j$ is defined as $I_p = \min_{(k,l) \in E(P)} w_{kl}$, where $E(P)$ is a set of edges along path $P$ and $w_{kl}$ is the weight of edge $(k, l)$ of path $P$, the value of which is defined in terms of the linguistic variables.

The general influence $Inf_{km}(i,j)$ of vertex $i$ on vertex $j$ is defined as follows: $Inf_{km}(i,j) = \max_{P(i,j)} I_p$, where max is the maximum value along all possible paths from vertex $i$ to vertex $j$. Thus, $I_p$ defines the weakest link in path $P$, and $Inf_{km}(i,j)$ defines the strongest influence among the indirect influences $I_p$.

## 3     Criteria of optimality of impact

Considering each possible simple path from node $u_i$ to the node $u_j$ of cognitive map as a certain scenario of impact $(u_i, u_j)_k$, it is necessity to determine criteria for choosing one of them.

The article presents two criteria for optimality of impact $C_1$ and $C_2$, which are called the force of impact and speed of implementation of the scenario respectively.

The force of impact of node $u_i$ on node $u_j$ is calculated for every path while considering the weights of the edges. The impulse from node $u_i$ is distributed along the path in the direction from $u_i$ to $u_j$ according to rules a)–d) [5]:

**a)** $u_i \xrightarrow{+} u_k \xrightarrow{-} u_j$

If node $u_i$ has a positive impact on node $u_k$ and node $u_k$ has a negative impact on node $u_j$, then node $u_i$ is said to increase the negative impact of $u_k$ on $u_j$. As a result, node $u_i$ is said to have a negative impact on $u_j$.

**b)** $u_i \xrightarrow{-} u_k \xrightarrow{-} u_j$

If node $u_i$ decreases the negative impact of node $u_k$ on $u_j$, then node $u_i$ is said to have a positive impact on $u_j$.

**c)** $u_i \xrightarrow{+} u_k \xrightarrow{+} u_j$

In this case, $u_i$ has a positive impact on $u_j$, which increases the positive impact of node $u_k$ on $u_j$.

**d)** $u_i \xrightarrow{-} u_k \xrightarrow{+} u_j$

In this case, node $u_i$ has a negative impact on node $u_k$ and $u_k$ has a positive impact on $u_j$. In other words, node $u_i$ decreases the positive impact of $u_k$ on $u_j$. Thus, node $u_i$ has a negative impact on $u_j$.

The full impact $z_{ij}$ on the node $u_j$, which is accumulated from the node $u_i$, is the sum of the partial impacts calculated as subtract between $z_{ij}^k - \tilde{z}_{ij}^k$ in all simple paths from node $u_i$ to node $u_j$ (following to the algorithm for calculating of a mutual impact between nodes in weighted graphs – the algorithm of an accumulative impact, which is presented in [5])
where

$$z_{ij}^k(t+1) = \left(1 + \text{sign}(z_{ij}^k(t)) * \alpha\left(\left|\frac{z_{ij}^k(t)}{\mu}\right|\right)\right) * w(q_t^k, q_{t+1}^k),$$

$$\tilde{z}_{ij}^k(r+1) = \left(1 + \text{sign}(\tilde{z}_{ij}^k(r)) * \alpha\left(\left|\frac{\tilde{z}_{ij}^k(r)}{\mu}\right|\right)\right) * w(q_r^k, q_{r+1}^k),$$

$q_t^k$ – the sequence of nodes included in the $k$-th path ($q_0 = u_i$, $q_{m-1} = u_j$);
$t = 0, 1, ..., m-2$, a $r = 1, ..., m-2$ ($m$ is the number of nodes included to the $k$-th

path).

Here, the initial conditions are: $z_{ij}^k(0) = 0$, $\tilde{z}_{ij}^k(1) = 0$.

$$\mu = \max |w_{ij}|,$$

where $i = 0,1,...,n$, $j = 0,1,...,n$ ($n$ is the dimension of the cognitive map).

The impact of a node $u_i$ on a node $u_j$ is called "the strongest", if the partial impact on the final node is characterized by the greatest of absolute magnitude of an impact among all of the partial impacts on all other simple paths between two nodes $u_i$ and $u_j$.

The impact of the node $u_i$ on node $u_j$ is considered to be "the fastest in realization", if it is carried out in the shortest path. The speed of the implementation of the $k$-th scenario is determined by the number of edges ($m-1$) connecting the nodes $u_i$ and $u_j$ in $k$-th path (where $m$ is the number of nodes included in the $k$-th path).

The introduced criteria of $C_1$ and $C_2$ are almost equivalent in terms of priority, thereby if one get several different optimal scenarios $(u_i, u_j)_k$ of the impact of the node, one cannot just select neither of them. Therefore, the fundamental complexity of choice in multi-criteria problems consists in impossibility of determining the optimal scenario a priori. So, there is a need to compare alternatives to all criteria.

Let us consider $X$ as a set of possible scenarios (alternatives) $(u_i, u_j)_k$ of the impact of the node $u_i$ on $u_j$. The minimum number of elements included in the set $X$ is two (to be able to make a choice). There is no limit on the number of possible scenarios: the number of elements of the set can be both finite and infinite. It is worth noting that sometimes a choice of not one, but an entire set of decisions is made, which is a subset of a set of possible solutions $X$. In this article, it is necessary to justify the choice of the optimal scenario of impact according to the introduced criteria $C_1$ and $C_2$. Then $C(X)$ is a set of selected scenario. It is a solution of the problem of choice and it can be any subset of the set of possible scenarios $X$. Thus, solving the problem of choice means to find a subset of $C(X)$, $C(X) \subset X$.

In the case where a plurality of selected scenarios does not contain any element, the choice does not occur, due to the fact that no solution has been selected. That is, in order to make the choice, it is necessary that the set $C(X)$ contains at least one element.

There are various methods for solving multi-criteria problem [9]. In order to obtain a unified solution to the problem posed in this article, a multi-criteria assessment of the scenarios obtained according to the Pareto principle [10], [11], [12] is used.

Pareto's approach is as follows: the alternative is "the best" than the alternative for Pareto ($x \succ y$), if alternative $x$ alternatives are rated "no worse" than alternatives $y$, and at least one alternative $x$ is "the best" than alternatives $y$:

$$\forall_i \ C_i(x) \geq C_i(y) \text{ and } \exists_j : C_j(x) > C_j(y)$$

where $C(X)$ is a function of choice ($C(X) \subset X$).

The resulting set of solutions is called pareto-optimal.

## 4 Method for Searching of an Optimal Scenario of Impact

Let us set of possible vectors $X$ consist of a finite number of elements $N$ and has the form $X = \{x^{(1)}, x^{(2)}, ..., x^{(N)}\}$.

In order to construct it on the basis of the definition of the Pareto set, it is necessary to compare each vector $x^{(i)} \in X$ with any other vector $x^{(j)} \in X$. Thereby, a step-by-step comparison of scenarios (corresponding columns of the table) based on the principle of "no less" ("no more") according to all criteria is performed. Namely: if the *i*-th scenario is larger (at minimization) or smaller (at maximization) of $j$-th scenario by at least one criterion, then this scenario is no longer taken into account. But if at least one $i$-th scenario criterion is less (at minimization) or larger (at maximization) for $j$-th scenario, with one or more other criteria, it is greater (at minimization) or smaller (at maximization), then both scenarios are taken into account.

It must be pointed out that it is convenient to use a table whose rows are criteria $C_1$ and $C_2$ (a force of impact and ease of implementation of the scenario, respectively), and the columns are the number of a scenario $(u_i, u_j)_k$ (the numbers of simple paths connecting the nodes $u_i$ and $u_j$) for comparison alternatives.

Thus, columns of a table form a set of possible vectors (possible scenarios), which consist of two elements - the values of the criteria. The result of a staged comparison is the set $C(X)$ of such non-extracted vectors forms the Pareto set. But often this is the case, and as already mentioned above, the Pareto set may contain more than one element. These are scenarios that cannot be compared according to the Pareto principle. In the general case, when the Pareto set contains more than one element, in order to determine the optimal scenario of impact in this article, the following algorithm is proposed:

a) Firstly, the least common multiple (LCM) of the criterion $C_2$ for all values of the Pareto set is determined. Considering $C_2$ as time the corresponding scenario is implemented for, then the LCM of all values is the least time for which the integer number of each of the scenarios included in the Pareto set is realized. Thereby, at the same time $LCM(c_2^{(1)}, ..., c_2^{(d)})$, the number of realizations of the various scenarios included in the Pareto set are different accordingly $\{a^{(1)}, a^{(2)}, ..., a^{(d)}\}$:

$$a^{(k)} = \frac{LCM(c_2^{(1)},...,c_2^{(d)})}{c_2^{(k)}}$$

where $c_2^{(k)}$ – value of the criterion $C_2$ for $k$-th scenario;

$d$ – the number of elements included in the Pareto set.

b) Next, for each of the scenarios included in the Pareto set, the values of their assessments by the criterion $C_1$ $\{c_1^{(1)}, c_1^{(2)},...,c_1^{(d)}\}$ are multiplied by the corresponding value $\{a^{(1)}, a^{(2)},...,a^{(d)}\}$. That is, it determines what will be the overall impact of the node $u_i$ on node $u_j$ the time $LCM(c_2^{(1)},...,c_2^{(d)})$ of the $k$-th scenario.

c) In order to determine the optimal scenario of impact, it is necessary to find the highest value of the multiplication $c_1^{(k)} * a^{(k)}$ defined in step b):

$$\max_{k} = c_1^{(k)} * a^{(k)}$$

where $k = 1,..,d$.

That is, the number $k$ to which the largest multiplication $c_1^{(k)} * a^{(k)}$ corresponds is the number of the optimal scenarios of impact. As a result of the justification of the choice of the optimal criteria $C_1$ and $C_2$ the impact scenario for each pair of nodes $(u_i, u_j)$ of a weighted graph, we can construct a matrix $Z$ which consists of elements $z_{ij}$ and a matrix $T$ which consists of elements $t_{ij}$.

*def 1*: The full impact $z_{ij}$ is the partial impact of the node $u_i$ on the node $u_j$, which is accumulated in accordance with the optimal scenario of impact (i.e., the value of the criterion $C_1$ of the optimal scenario of impact). If $u_j$ is unavailable from node $u_i$, then $z_{ij} = 0$.

*def 2*: The full time $t_{ij}$ is the time it takes to implement the optimal scenario of impact node $u_i$ on node $u_j$ (i.e., the value of the criterion $C_2$ of the optimal scenario of impact). If $u_j$ is unreachable from the node $u_i$, then $t_{ij} = 0$.

In order to determine what the impact of each of the nodes at $t \to \infty$, must be fulfilled as follows. Firstly $z_{ij}^1$ needed to be defined $z_{ij}^1$ - the impact of each node at $t = 1$. Taking into account the time required to implement each of the scenarios for the impact matrix $Z$ by dividing each of its non-zero elements $z_{ij}$ ($z_{ij} \neq 0$) into the corresponding element $t_{ij}$ of the matrix $T$, a matrix $Z_1$ is obtained, the elements of which are:

$$z_{ij}^1 = \begin{cases} \dfrac{z_{ij}}{t_{ij}}, & z_{ij} \neq 0 \\ 0, & z_{ij} = 0 \end{cases}.$$

Next, the impact of each node in time $t$ is calculated as $z_{ij}^{t+1} = z_{ij}^t + z_{ij}^1$.

At each step of $t = 2, 3, 4...$, the process of normalization is carried out:

$$z_{ij}^t = \dfrac{z_{ij}^t}{\sum_{k=1}^{n} \sum_{l=1}^{n} z_{kl}^t}.$$

## 5   Example

The work [12] considered the weighted directed graph shown in Fig. 1.

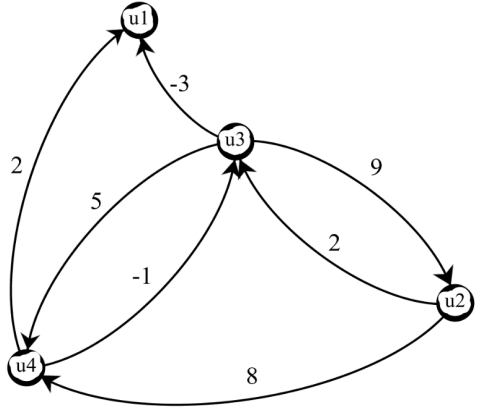

**Fig. 1.** Weighted directed graph.

The weighted directed graph, presented in Fig. 1, corresponded to cognitive map is defined by the adjacency matrix:

$$W = \begin{pmatrix} 0 & 0 & 0 & 0 \\ 0 & 0 & 2 & 8 \\ -3 & 9 & 0 & 5 \\ 2 & 0 & -1 & 0 \end{pmatrix} \qquad (1)$$

Table 1 demonstrates an example of assessment of impact scenario of nodes $u_3$ on nodes $u_4$ by criteria $C_1$ and $C_2$:

**Table 1.** Example of Assessment of Impact Scenario of Nodes

| $(u_1, u_4)_k$ | Siple path from $u_1$ to $u_4$ | $C_1$ | $C_2$ |
|---|---|---|---|
| 1 | $u_3 \xrightarrow{5} u_4$ | 6,92 | 2 |
| 2 | $u_3 \xrightarrow{9} u_2 \xrightarrow{8} u_4$ | 5 | 1 |

Table 2 shows the Pareto table to find the optimal criteria $C_1$ and $C_2$ scenario of the node $u_3$ impact on node $u_4$ for the cognitive map, which is shown in Fig. 1.

**Table 2.** Pareto Table to Find the Optimal Criteria

| $(u_3, u_4)_k$ | 1 | 2 |
|---|---|---|
| $C_1$ | 6,92 | 5 |
| $C_2$ | 2 | 1 |

In this case, the Pareto set consists of two non-comparable vectors (two scenarios 1 and 2), among which it is impossible to determine uniquely optimal by criteria $C_1$ and $C_2$ (scenario 1 is the optimal by criterion $C_1$, and scenario 2 is by $C_2$ one). Therefore, for the final solution of the problem of choosing the optimal scenario of impact, it is necessary to determine the alternative to the optimal solution for a particular practical problem.

According to the method for searching of an optimal scenario of impact, which is proposed in this article, when the Pareto set contains more than one element, it is first necessary to find $LCM$ of values of the criterion $C_2$ values for all elements of the Pareto set. For the Pareto set constructed from the set of alternatives presented in Table 2, the least time for which the integer number of each of the scenarios included in this Pareto set is equal:

$$LCM(2,1) = 2.$$

The number of implementations of the first and second scenarios will be equal respectively

$$a^{(1)} = \frac{LCM(2,1)}{2} = \frac{2}{2} = 1,$$

$$a^{(2)} = \frac{LCM(2,1)}{1} = 2.$$

Over time equal to $LCM(2,1) = 2$, the overall impact of the node $u_3$ on node $u_4$ the 1st scenario $(u_3, u_4)_1$ is:

$$c_1^{(1)} * a^{(1)} = 6.92 * 1 = 6.92$$

In the 2nd scenario

$$c_1^{(2)} * a^{(2)} = 5 * 2 = 10.$$

$$\max(c_1^{(1)} * a^{(1)}, c_1^{(2)} * a^{(2)}) = \max(6.92, 10) = 10$$

Therefore, for this example (Table 2), scenario number 2 $(u_3, u_4)_2$, in accordance with the proposed method, is optimal.

The Pareto table for other pairs of nodes $(u_i, u_j)$ $(i, j = 1, 2, 3, 4)$ with more than one scenario of impact is given in Table. 3.

**Table 3.** Pareto Table for Other Pairs of Nodes

| $(u_2, u_1)_k$ | 1 | 2 | 3 | 4 |
|---|---|---|---|---|
| $C_1$ | -1,08 | 0,41 | 1,66 | 0,22 |
| $C_2$ | 2 | 3 | 2 | 3 |
| $(u_2, u_3)_k$ | 1 | 2 | | |
| $C_1$ | 2 | -0,83 | | |
| $C_2$ | 1 | 2 | | |
| $(u_2, u_4)_k$ | 1 | 2 | | |
| $C_1$ | 8 | 1,79 | | |
| $C_2$ | 1 | 2 | | |
| $(u_3, u_1)_k$ | 1 | 2 | 3 | |
| $C_1$ | -3 | 0,27 | 1,34 | |
| $C_2$ | 1 | 3 | 2 | |
| $(u_4, u_1)_k$ | 1 | 2 | | |
| $C_1$ | 0,59 | 2 | | |
| $C_2$ | 2 | 1 | | |

For all possible scenarios presented in Table 3 for pairs $(u_2, u_1)$, the Pareto set will consist of one element $(u_2, u_1)_3$ - from scenario 3. For a pair $(u_2, u_3)$, the Pareto set consists of the element $(u_2, u_3)_1$ - scenario 1. For - $(u_2, u_4) - (u_2, u_4)_1$, for $(u_3, u_1) - (u_3, u_1)_1$ - and for the pair -. $(u_4, u_1) - (u_4, u_1)_2$.

After choosing the best scenario of impact by the introduced criteria $C_1$ and $C_2$ for each pair of nodes $(u_i, u_j)$ $(i, j = 1, 2, 3, 4)$ of the weighted directed graph represented in Fig. 1, we can construct an influence matrix $Z$ which consists of elements $z_{ij}$ (see *def 1*) and a matrix $T$ which consists of elements (see *def 2*):

$$Z = \begin{pmatrix} 0 & 0 & 0 & 0 \\ 1,66 & 0 & 2 & 8 \\ -3 & 9 & 0 & 5 \\ 2 & -1,79 & -1 & 0 \end{pmatrix}$$

$$T = \begin{pmatrix} 0 & 0 & 0 & 0 \\ 2 & 0 & 1 & 1 \\ 1 & 1 & 0 & 1 \\ 1 & 2 & 1 & 0 \end{pmatrix}$$

Taking into account the process of normalization at each step with $t \to \infty$, the impact of each of the nodes to other for the influence matrix $Z$ is represented in the form of the influence matrix:

$$Z_t = \begin{pmatrix} 0 & 0 & 0 & 0 \\ 0,038 & 0 & 0,091 & 0,365 \\ -0,137 & 0,41 & 0 & 0,228 \\ 0,091 & -0,04 & -0,046 & 0 \end{pmatrix} \quad (2)$$

The full impact $Inf_{pm}$ of each node for the influence matrix $Z_t$ is determined by the rule:

$$Inf_{pm}^i = \sum_{j=1}^n |z_{ij}^t| \quad (3)$$

where $n$ is the number of nodes of a cognitive map; $pm$ is a short for "the Pareto method".

The full impact $Inf_{pm}$ of each node and its rank distribution for the influence matrix (2), according to (3), are presented in Table 4.

**Table 4.** Rank Distribution of Nodes

| Node (№) | $Inf_{pm}$ |
|---|---|
| 3 | 0,775 |
| 2 | 0,494 |
| 4 | 0,178 |
| 1 | 0 |

Comparing the results of using the method for searching of an optimal scenario of impact with the results provided by the Kosko model for the adjacency matrix (1), it can be notice (Table 5) that the rank distribution of nodes by degree of impact, as a result of applying of each method, is remined.

**Table 5.** Rank Distribution of Nodes According to Kosko Model and Proposed Method

| Node (#) | $Inf_{km}$ | Node (#) | $Inf_{pm}$ |
|---|---|---|---|
| 3 | 20 | 3 | 0,775 |
| 2 | 12 | 2 | 0,494 |
| 4 | 4 | 4 | 0,178 |
| 1 | 0 | 1 | 0 |

In table 5 km is a short for "the Kosko model".

## 6    Conclusions

Consequently, the multi-criteria choice problem was considered in the article. Based on the criteria of a force of impact and speed of the implementation of the scenario, the choice of the optimal scenario of impact was justified. A comparison of the results of applying the method for searching of an optimal scenario of impact according to the introduced criteria, with the results which are obtained with applying the Kosko model was fulfilled.
   Using the results of these calculation, decision makers can develop strategic and tactical steps to counter-act the information operation, evaluate the operation`s efficiency.

## Acknowledgment

This study is funded by the NATO SPS Project CyRADARS (Cyber Rapid Analysis for Defense Awareness of Real-time Situation), Project SPS G5286.